\documentclass[letterpaper, 10 pt, conference]{ieeeconf}  % Comment this line out if you need a4paper

\IEEEoverridecommandlockouts                              % This command is only needed if 
\let\labelindent\relax

\usepackage{multicol}

\makeatletter
\let\NAT@parse\undefined
\makeatother

	\usepackage{amsmath}
	\usepackage{verbatim} % multi-line comments
	\usepackage{amsfonts}
	\usepackage{amssymb}
	\usepackage{ragged2e}
	\usepackage{graphicx}
	\usepackage{cancel}
	\usepackage{mathtools}
	\usepackage{tabularx}
     \usepackage{adjustbox} % vertical alignment of figures
	\usepackage{arydshln}
	\usepackage{tensor}
	\usepackage{array}
	\usepackage[dvipsnames]{xcolor}
	\usepackage{listings}
	\usepackage{textcomp}
	\usepackage{mathrsfs}
	\usepackage{bbm}
	\usepackage{tikz}
	\usepackage{tikz-cd}
	\usepackage{enumitem}
	\usepackage{arydshln}
	\usepackage{relsize}
	\usepackage{multirow}
	\usepackage{scalerel}
	\usepackage{upgreek}
	\usepackage{ifthen}
	\usepackage{yhmath}
	\usepackage{blkarray}
	\usepackage{dashrule}
	\usepackage{subcaption}
	\usepackage{overpic}
    \usepackage[commandnameprefix=ifneeded]{changes}

		\newcommand{\set}[1]{\left\{#1\right\}}

		\newcommand{\reals}{\mathbb{R}}
		\newcommand{\R}{\reals}

		\DeclareMathOperator{\trace}{tr}

            \DeclareMathOperator*{\minimize}{minimize}
		\DeclareMathOperator{\subto}{subject\,to}

	\usepackage{letltxmacro}
	\LetLtxMacro\orgvdots\vdots
	\LetLtxMacro\orgddots\ddots

	\makeatletter
	\DeclareRobustCommand\vdots{%
		\mathpalette\@vdots{}%
	}
	\newcommand*{\@vdots}[2]{%
		\sbox0{$#1\cdotp\cdotp\cdotp\m@th$}%
		\sbox2{$#1.\m@th$}%
		\vbox{%
			\dimen@=\wd0 %
			\advance\dimen@ -3\ht2 %
			\kern.5\dimen@
			\dimen@=\wd2 %
			\advance\dimen@ -\ht2 %
			\dimen2=\wd0 %
			\advance\dimen2 -\dimen@
			\vbox to \dimen2{%
				\offinterlineskip
				\copy2 \vfill\copy2 \vfill\copy2 %
			}%
		}%
	}
	\DeclareRobustCommand\ddots{%
		\mathinner{%
			\mathpalette\@ddots{}%
			\mkern\thinmuskip
		}%
	}
	\newcommand*{\@ddots}[2]{%
		\sbox0{$#1\cdotp\cdotp\cdotp\m@th$}%
		\sbox2{$#1.\m@th$}%
		\vbox{%
			\dimen@=\wd0 %
			\advance\dimen@ -3\ht2 %
			\kern.5\dimen@
			\dimen@=\wd2 %
			\advance\dimen@ -\ht2 %
			\dimen2=\wd0 %
			\advance\dimen2 -\dimen@
			\vbox to \dimen2{%
				\offinterlineskip
				\hbox{$#1\mathpunct{.}\m@th$}%
				\vfill
				\hbox{$#1\mathpunct{\kern\wd2}\mathpunct{.}\m@th$}%
				\vfill
				\hbox{$#1\mathpunct{\kern\wd2}\mathpunct{\kern\wd2}\mathpunct{.}\m@th$}%
			}%
		}%
	}
	\makeatother

	\tikzset{
	  symbol/.style={
		draw=none,
		every to/.append style={
		  edge node={node [sloped, allow upside down, auto=false]{$#1$}}}
	  }
	}
\usepackage[bookmarks=true]{hyperref}

\usepackage{cleveref}
\Crefname{figure}{Figure}{Figures}
\Crefname{table}{Table}{Tables}
\Crefname{equation}{Eq.}{Eqs.}
\Crefname{section}{Section}{Sections}
\Crefname{subsection}{Subsection}{Subsections}
\Crefname{appendix}{Appendix}{Appendices}

\usepackage[commentColor=black]{algpseudocodex}
\tikzset{algpxIndentLine/.style={draw=black}}
\algrenewcommand{\alglinenumber}[1]{\bfseries\footnotesize #1}
\algrenewcommand{\textproc}{}
\algrenewcommand{\algorithmicrequire}{\textbf{Input:}}
\algrenewcommand{\algorithmicensure}{\textbf{Output:}}
\usepackage{algorithm}
\makeatletter
\newcommand{\algorithmname}{\ALG@name}
\renewcommand{\floatc@ruled}[2]{{\@fs@cfont #1:} #2\par}
\makeatother

\usetikzlibrary{shapes.geometric}
\usetikzlibrary{angles, quotes, calc}
\usetikzlibrary{arrows.meta}

\usepackage[acronym]{glossaries}

\newacronym{gcs}{GCS}{Graph of Convex Sets}
\newacronym{qcqp}{QCQP}{Quadratically Constrained Quadratic Program}
\newacronym{qcqps}{QCQPs}{Quadratically Constrained Quadratic Programs}
\newacronym{sdp}{SDP}{Semidefinite Programming}
\newacronym{sdr}{SDR}{Semidefinite Relaxation}
\newacronym{sdrs}{SDRs}{Semidefinite Relaxations}
\newacronym{mip}{MIP}{Mixed-Integer Programming}
\newacronym{micp}{MICP}{Mixed-Integer Convex Program}
\newacronym{spp}{SPP}{Shortest Path Problem}
\newacronym{com}{COM}{Center of Mass}
\newacronym{soc}{SOC}{Second-Order Cone}
\newacronym{mpc}{MPC}{Model-Predictive Controller}
\newacronym{lp}{LP}{Linear Program}
\newacronym{sdf}{SDF}{Signed Distance Function}
\newacronym{socp}{SOCP}{Second-Order Cone Program}
\newacronym{rsoc}{RSOC}{Rotated Second-Order Cone}
\newacronym{rlt}{RLT}{Reformulation-Linearization Technique}
\newacronym{nlp}{NLP}{Nonlinear Program}
\newacronym{nlps}{NLPs}{Nonlinear Programs}
\newacronym{bb}{BB}{Branch-and-Bound}
\newacronym{rl}{RL}{Reinforcement Learning}
\newacronym{bc}{BC}{Behavior Cloning}
\newacronym{pwa}{PWA}{Piecewise Affine}
\newacronym{psd}{PSD}{Positive Semi-Definite}
\newacronym{mincp}{MINCP}{Mixed-Integer Nonconvex Programming}
\newacronym{sos}{SOS}{Sums-of-Squares}
\usepackage{color-edits}

\addauthor{bpg}{cyan}
\addauthor{rt}{orange}
\addauthor{tm}{magenta}
\addauthor{pp}{red}
\addauthor{aa}{purple}
\addauthor{sy}{green}
\addauthor{sm}{brown}

\title{\LARGE \bf
Towards Tight Convex Relaxations for Contact-Rich Manipulation
}

\author{
Bernhard Paus Graesdal$^1$,
Shao Yuan Chew Chia$^2$,
Tobia Marcucci$^1$, \\
Savva Morozov$^1$,
Alexandre Amice$^1$,
Pablo A. Parrilo$^1$, and
Russ Tedrake$^1$% <- this % stops a space
\thanks{$^1$ Department of Electrical Engineering and Computer Science, Massachusetts Institute of Technology, Cambridge, MA, USA.
E-mail: \texttt{\{graesdal, tobiam, savva, parrilo, russt\}@mit.edu}}
\thanks{
$^2$ Department of Computer Science, Harvard University, Cambridge MA, USA. E-mail: \texttt{shaoyuan\_chewchia@college.harvard.edu}
}
}

\begin{document}

\maketitle
\thispagestyle{empty}
\pagestyle{empty}
%\tmcomment{Title: maybe mention ``global''? Personally I don't love the ``using...'' in a title, since I think the title should be more about the what rather than the how. Also I'm not sure if we want to frame this paper around planar pushing, or more use it as a toy example of control through contact.}

\begin{abstract}
%!TEX root = ../root.tex
We present a novel method for global motion planning of robotic systems that interact with the environment through contacts. Our method directly handles the hybrid nature of such tasks using tools from convex optimization.
We formulate the motion-planning problem as a shortest-path problem in a graph of convex sets, where a path in the graph corresponds to a contact sequence and a convex set models the quasi-static dynamics within a fixed contact mode.
For each contact mode, we use semidefinite programming to relax the nonconvex dynamics that results from the simultaneous optimization of the object's pose, contact locations, and contact forces.
The result is a tight convex relaxation of the overall planning problem, that can be efficiently solved and quickly rounded to find a feasible contact-rich trajectory.
As an initial application for evaluating our method, we apply it on the task of planar pushing.
Exhaustive experiments show that our convex-optimization method generates plans that are consistently within a small percentage of the global optimum, without relying on an initial guess, and that our method succeeds in finding trajectories where a state-of-the-art baseline for contact-rich planning usually fails.
We demonstrate the quality of these plans on a real robotic system.

%Motion planning for planar pushing is a hybrid problem, where one needs to plan contact locations and what forces to apply, while considering friction, rigid body dynamics, and contact modes.
%We present a novel motion planning framework which is able to address both the combinatorial and continuous aspect of this problem simultaneously.
%We formulate the hybrid motion planning problem as a shortest path problem in a graph of convex sets, where a path in the graph corresponds to a mode sequence and the convex sets correspond to contact manifolds. Using tools from convex optimization, 
%Planar pushing is a hybrid problem that requires reasoning about contact modes, dynamics, and friction. 
%In contrast to prior works, our framework is able to address both the full combinatorial and continuous aspect of the motion planning problem simultaneously.
\end{abstract}

\section{Introduction}
\label{sec:introduction}
%!TEX root = ../root.tex

% Optimal planning and control through contact is an important and challenging problem in robotics, and includes tasks like robot locomotion and hybrid and underactuated dynamical system, making planning and control difficult.

Optimal planning and control through contact is an important and challenging problem in robotics, which includes tasks like robot locomotion and dexterous manipulation.
%Its key challenge is the hybrid nature of the contact dynamics, which requires reasoning about discrete contact modes and continuous trajectories simultaneously.
It generally involves both a hybrid and underactuated dynamical system, making planning and control difficult.
A variety of methods have been proposed to tackle this problem.
Many of them artificially decouple the discrete and the continuous components of the problem, yielding suboptimal trajectories that do not take full advantage of the rich contact dynamics.
Approaches that blend the discrete and continuous components often do so locally (around a given trajectory) and are unable to reason about the global problem; or rely on expensive global optimization algorithms that scale poorly.

\begin{figure}
    \centering    
    \includegraphics[width=0.95\linewidth]{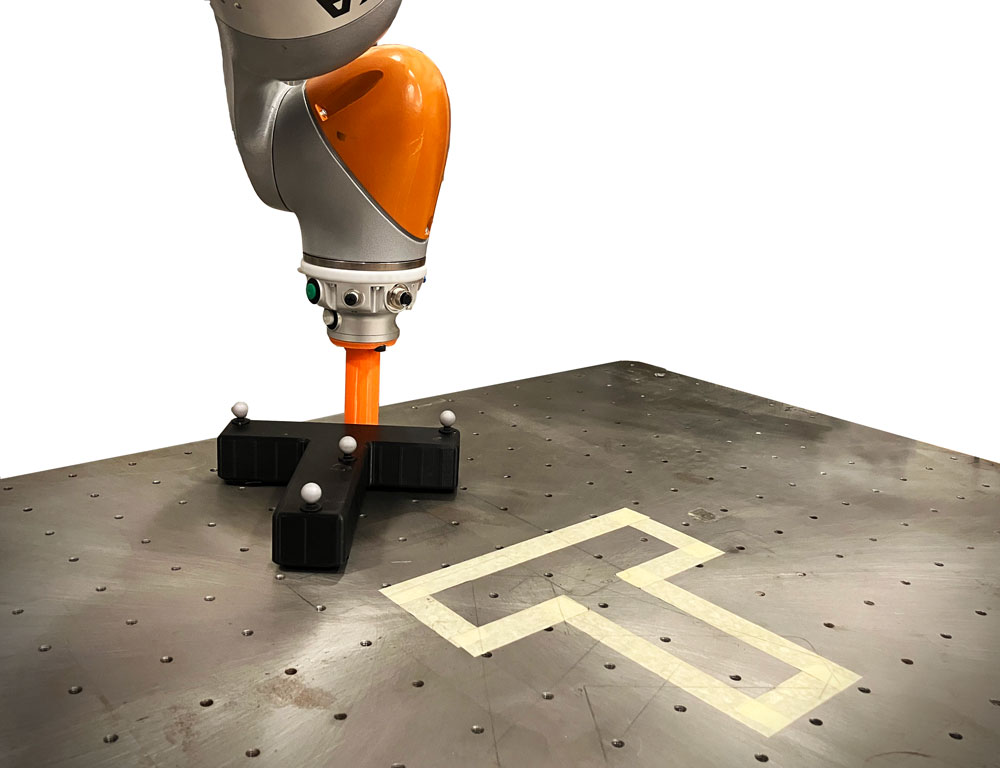}
    \caption{The experimental planar pushing setup. A cylindrical finger is attached to a robotic arm that is pushing a T-shaped object on the table into its target configuration.}
    \label{fig:experimental_setup}
\end{figure}

In this work, we introduce a method that naturally blends the discrete logic and the continuous dynamics of planning through contact into a convex optimization problem.
Specifically, we formulate the planning problem as a \acrfull*{spp} in a \acrfull*{gcs}, a class of mixed discrete/continuous optimization problems that can be effectively solved to global optimality~\cite{marcucci2023shortest}.
Paths in this graph correspond to different contact sequences, and the convex sets model the quasi-static contact dynamics within a fixed contact mode.
These quasi-static dynamics are bilinear (therefore nonconvex), since we simultaneously optimize over the object pose, contact locations, and contact forces.
Our method approximates these bilinearities using a tight \acrfull*{sdp} relaxation for each contact mode.
The methods from~\cite{marcucci2023shortest} are then leveraged to produce a tight convex relaxation of the global planning-through-contact problem.
This relaxation can be quickly solved and rounded to obtain a feasible contact-rich trajectory.
By comparing the cost of the rounded solution and the convex relaxation, our motion planner also provides us with tight optimality bounds for the trajectories that it designs.

As a first application for evaluating our method,
this work explores the task of planar pushing, first studied by Mason in~\cite{masonMechanicsPlanningManipulator1986}.
Planar pushing has applications that span from warehouse automation to service robotics, and although it is among the simplest examples of non-prehensile manipulation, current state-of-the-art approaches are still unable to solve this problem reliably to global optimality.
The method proposed in this paper can simultaneously reason about the high-level discrete mode switches and the low-level continuous dynamics of planar pushing in a global fashion.
We evaluate our motion planner through thorough numerical experiments, which show that the trajectories we generate typically have a very small optimality gap ($10\%$ on average).
We also demonstrate our approach on the real robotic system shown in in~\Cref{fig:experimental_setup}.
We emphasize that, although we study the tightness of our relaxations on the particular domain of planar pushing, the technique we introduce generalizes naturally to more complex multi-contact problems.

\section{Related Work}
\label{sec:related_work}
%!TEX root = ../root.tex

In this section we review the multiple approaches that have been proposed in the literature to solve planning problems involving contacts.
We categorize the existing approaches into learning-based, sampling, local and global optimization.

The learning methods that have been proven most effective for contact-rich tasks are \acrfull*{rl} and \acrfull*{bc}.
%In particular, \acrshort*{rl} has shown impressive robustness in quadruped locomotion~\cite{hwangbo2019learning, lee2020learning, siekmann2021sim}, and has been successfully applied to manipulation tasks in simulation~\cite{lillicrap2015continuous,rajeswaran2017learning}, real-world robotic manipulation~\cite{gu2017deep, pmlr-v87-kalashnikov18a}, and more recently 
\acrshort*{rl} has proven particularly effective due to its ability to bypass the low-level combinatorial contact decisions by using stochastic approximation of the system's dynamics~\cite{suh2022differentiable, suh2022bundled, lidec2022leveraging}.
In particular, \acrshort*{rl} has shown impressive robustness for real-world quadruped locomotion~\cite{hwangbo2019learning}, and has been successfully applied to manipulation tasks in simulation~\cite{lillicrap2015continuous,rajeswaran2017learning}, real-world robotic manipulation~\cite{gu2017deep, pmlr-v87-kalashnikov18a}, and more recently 
on real-world dexterous hands \cite{andrychowicz2020learning, chen2022system, qi2023general, yin2023rotating}.
%\acrshort*{rl} has proven particularly effective for contact problems because it abstracts away low-level combinatorial decision making via sampling and averaging.
% One reason why RL through contact has proven particularly effective is because it abstracts low-level combinatorial decision making via sampling and averaging.
%\acrshort*{rl} has proven particularly effective due to its ability to bypass the process of making low-level combinatorial contact decisions by using stochastic approximation schemes \cite{suh2022differentiable, suh2022bundled, lidec2022leveraging} \bpgcomment{TODO: Simplify this sentence}.
% \aacomment{I don't know what this means.}
% \smcomment{i'm referencing Terry's work here to imply "RL does low-level better than opt-based-methods". will talk to him about how to phrase it better.}
% , but struggle with general long-horizon planning.
% (with recent methods attempting to address these issues\cite{ma2023eureka} \smcomment{refs}).
Similarly, \acrshort*{bc} methods have been successfully applied to many real-world manipulation tasks~\cite{chi2023diffusion, octo_2023, shafiullah2023bringing, janner2022planning}, including planar pushing~\cite{chi2023diffusion, florence2022implicit}.
% planning and navigation~\cite{janner2022planning, sridhar2023nomad}
% \bpgcomment{Seeing as these are not related to contact-rich tasks or planar pushing, should we remove/replace these?}, 
Although \acrshort*{rl} and \acrshort*{bc} are applicable to a variety of systems, \acrshort*{rl} methods tend to require expensive training and task-specific user-shaped reward functions, while \acrshort*{bc} requires costly expert demonstrations that are typically generated by human teleoperation.
The method we propose in this paper can potentially be used to automatically generate large sets of expert demonstrations of very high quality for a variety of different tasks.

Sampling-based methods have also proven effective for planning contact-rich trajectories.
These algorithms often alternate between a high-level discrete search and a low-level continuous trajectory optimization.
Although, in principle, this facilitates some level of global reasoning, the inability to optimize the discrete and continuous decisions jointly makes it hard to find globally optimal solutions, or even quantify the suboptimality of the solutions found.
Another key challenge is that the feasible robot configurations lie on lower-dimensional contact manifolds, necessitating guided exploration in sampling.
Existing methods either assume the ability to sample from individual contact modes~\cite{chengContactModeGuided2021a, chengContactModeGuided2022, vega2020asymptotically,chen2021trajectotree}, guide the search by estimating reachable sets in configuration space~\cite{chavan-dafle2020planar, wu2020r3t}, 
or combine smoothed contact models with task-specific contact samplers~\cite{pangGlobalPlanningContactRich2022}.
The necessity for hand-crafted samplers, the lack of global reasoning, and the uncertainty around the plan quality often makes sampling-based contact planners difficult to use in practice. 

Local optimization methods typically use contact-implicit models, where the contact is modelled as the solution set of complementarity constraints. 
These methods are amenable to continuous optimization by either producing smooth approximations to the contact dynamics~\cite{mordatch2012discovery, tassa2012synthesis, tassa2014control, chatzinikolaidis2021trajectory, howell2022trajectory}, or directly operating on the complementarity formulation~\cite{posaDirectMethodTrajectory2014, sleiman2019contact, manchester2020variational, jinTaskDrivenHybridModel2022, wangContactImplicitPlanningControl2023}. 
While these methods have shown great results in both locomotion and manipulation, they are inherently local and require high-quality initial guesses, which can be difficult to obtain \cite{wensing2023optimization}.

Global optimization methods for planning through contact typically use \acrfull*{mip} to jointly optimize over discrete contact modes and continuous robot motions.  
A common approach is to produce a \acrfull*{pwa} approximation of the nonlinear dynamics, casting the problem as an \acrshort*{mip}~\cite{marcucci2019mixed}.
This and similar approaches have been applied to push-recovery~\cite{marcucci2017approximate}, locomotion~\cite{aceituno2017simultaneous},
and feedback control for planar pushing~\cite{hoganReactivePlanarNonprehensile2020,hoganFeedbackControlPusherSlider2020}.
Recently, efficient task-specific \acrshort*{pwa} approximation have also been obtained by pruning large neural networks~\cite{liu2023model}.
The main limitation of these approaches is the \acrshort*{mip} runtime, which can grow exponentially with the number of integer variables.
In addition, \acrshort*{pwa} models are also unable to accurately capture the smooth contact dynamics within a fixed contact mode.
At the cost of a further slowing down computations, the latter issue can be tackled by introducing additional binary variables and approximating the nonconvexities with piecewise linear functions (see, e.g.,~\cite{deits2014footstep} or~\cite{aceituno-cabezasGlobalQuasiDynamicModel2020}), possibly in combination with tailored convex relaxations~\cite{ponton2016convex}.
Alternatively, general \acrfull*{mincp} approaches have been used in robotics for momentum-based footstep planning~\cite{koolenBalanceControlLocomotion2020}, but their applicability is limited to very small problem instances.

Our method is similar to the ones in the last category, but is substantially more efficient.
Rather than resorting to \acrshort*{mincp} or piecewise linear approximations combined with mixed-integer programming,
we depart from prior work by leveraging a tight convex relaxation for the bilinearities in each contact mode
while also leveraging the methods from~\cite{marcucci2023shortest} to produce a tight convex relaxation for the global planning-through-contact problem. 
The tightness of both relaxations enables us to obtain near-globally optimal solutions by solving the entire problem as a single \acrshort*{sdp}, followed by a quick rounding step.
This is the first work to our knowledge that is able to effectively solve the global planning-through-contact problem as a single convex program.

\section{Problem Statement}
\label{sec:problem_statement}
As a first application of our method, we explore planar pushing, a non-prehensile manipulation task where the robot uses a cylindrical finger to manipulate the motion of an object resting on a flat surface.
We shall refer to the manipulated object as the \textit{slider}, the robot finger as the \textit{pusher}, and to the overall system as the \textit{slider-pusher} system.
We assume that the slider and the pusher are rigid bodies, and restrict the problem to planar motion, such that the slider is always making contact with the underlying surface.
We also assume that the slider is a prism-shaped object with known geometry, which can be nonconvex, and a known (but arbitrary) \acrshort*{com}.
See \Cref{fig:experimental_setup} for a picture of the the real-world problem setup.

As the slider is pushed, it is subject to a contact wrench from the pusher and a friction wrench from the underlying surface (we assume a uniform pressure distribution between the slider and the surface).
We assume quasi-static dynamics (i.e., no work is done by impacts) and that the velocities are sufficiently low to make inertial forces negligible.
Under this assumption, for the slider to move, the contact wrench applied by the pusher must balance the friction wrench from the underlying surface.
We assume isotropic Coulomb friction, i.e., that the coefficient of friction is constant, and the friction force at every contact point must have a constant magnitude and oppose the direction of motion during sliding.
%\replaced{
Although these assumptions are common in the literature on non-prehensile modelling~\cite{goyalPlanarSlidingDry1991,lynchStablePushingMechanics1996} and planning \cite{doshiHybridDifferentialDynamic2020,hoganReactivePlanarManipulation2018}, we note that a more complex friction model could in principle be incorporated directly into our framework or as a post-processing step.
%}
%{
%These assumptions are standard in the non-prehensile literature~\cite{goyalPlanarSlidingDry1991,lynchStablePushingMechanics1996,hoganReactivePlanarManipulation2018}.
%}

The described slider-pusher system is a hybrid, underactuated system. 
The problem is underactuated because the slider cannot move on its own, and because Coloumb friction limits the range of contact forces that can be applied by the pusher.
Furthermore, the problem has hybrid dynamics, as there can be either no contact between the slider and pusher, or the contact can be sticking, sliding left, or sliding right relative to the contact point.

\section{High-Level Approach}
\label{sec:high_level_approach}
The first step in formulating our motion planning method is to consider the dynamics and kinematics in a fixed contact mode.
By formulating the manipulation problem in the task space, we are able to describe the system with only quadratic and linear constraints.
More specifically, quadratic equality constraints arise from considering elements of \(\mathrm{SO}(2)\), rotations of velocities and forces to the world frame, and the contact torque as the cross-product between the contact arm and force.
We thus formulate the motion planning problem within a specific contact mode as a nonconvex \acrfull*{qcqp}, which we approximate as a convex program using a semidefinite relaxation.

The second step in our method is to formulate the global motion planning problem as an \acrshort*{spp} in a \acrshort*{gcs}~\cite{marcucci2023shortest}.
For each of the contact modes in the slider-pusher system we add its associated convex program as a vertex to the graph, meaning that the feasible set of trajectories is added as a convex set and the cost is added as the vertex cost.
Additionally, we decompose the collision-free subset of the configuration space into convex regions, which are also added to the graph to encode collision-free motion planning, similar to \cite{marcucci2023motionplanning}.
By representing the desired initial and target configurations with vertices in the graph, the globally optimal solution to the motion planning problem now corresponds mathematically to the shortest path from the source to the target in the graph.

The techniques from~\cite{marcucci2023shortest} allow us to write a convex relaxation of this planning problem which is itself an \acrshort*{sdp} that can be solved efficiently. The solution of this relaxation is then rounded to obtain a feasible solution to the motion planning problem, where the optimality gap can be bounded by comparing the cost of the feasible solution and the relaxation.
% \aacomment{This might be hand-holding too much/redundant with the introduction.}
%\tmcomment{I would add a few sentences here to say that the techniques from~\cite{marcucciShortestPathsGraphs2022} allow us to write a convex relaxation of this plannig problem that is itself an SDP. The solution of this relaxation is then rounded to get an approximate solution of the planning problem. The optimality gap can be bounded by comparing the cost of the feasible solution and the relaxation.}

%\section{Planning through contact as a shortest path problem}
%\label{sec:contact_spp}
%\input{sections/contact_spp}

\section{Background and Optimization Tools}
\label{sec:technical_background}

\subsection{Shortest paths in graph of convex sets}
\label{sec:gcs}
%!TEX root = ../root.tex

In this section, we briefly review the formulation for the \acrshort*{spp} in a \acrshort*{gcs}. For further details, the reader is referred to~\cite{marcucci2023shortest}.
We define a \textit{\acrfull*{gcs}} as a directed graph $G = (\mathcal{V},\mathcal{E})$ with vertex set $\mathcal{V}$ and edge set $\mathcal{E} \subset \mathcal{V}^2$,
where each vertex $v \in \mathcal{V}$ is paired with a compact convex set $\mathcal{X}_v$ and a point $x_v$ contained in the set. Additionally, each vertex $v$ is associated with a vertex cost $l_v(x_v)$, a nonnegative convex function of the point $x_v \in \mathcal{X}_v$. Finally, the edges of the graph are associated to convex constraints of the form $(x_u, x_v) \in \mathcal{X}_e$ which couple the vertices in an edge $e$.

A path $p$ in the graph $G$ is defined as a sequence of distinct vertices that connects a source vertex $s \in \mathcal{V}$ to the target vertex $t \in \mathcal{V}$. 
Let $\mathcal{P}$ denote the family of all $s$-$t$ paths in the graph, and let $\mathcal{E}_p$ denote the set of edges traversed by the path $p$.
%\aacomment{
%\begin{definition}
%A \emph{Graph of Convex Sets} is a directed graph $G = (\mathcal{V},\mathcal{E})$ where each vertex consists of a tuple $v =(x_{v}, \mathcal{X}_{v}) \in \mathcal{V}$ where $\mathcal{X}_{v}$ is a bounded convex set, and $x_{v} \in \mathcal{X}_{v}$. The vertices of $G$ are weighted by non-negative, convex vertex costs $l_{v}(x_{v})$ and the edges are weighted by non-negative, convex edge costs
%\end{definition}
%The above isn't quite correct, but I think it would be good to package it into a definition.
%}
%\tmcomment{I prefer the definition that Bernhard gave. Although I would try to not make a hard stop at the end of the sentence ``We define a \acrshort*{gcs} as a directed graph $G = (\mathcal{V},\mathcal{E})$ with vertex set $\mathcal{V}$ and edge set $\mathcal{E} \subset \mathcal{V}^2$.'' Since if one extract this sentence from the context this is not quite true.}
%\aacomment{Fair enough. I just think that "GCS" and "SPP in a GCS" is used a lot, that it is useful if the reader can pick that out easily hence the definition environment.}
%\rtcomment{Along those lines, in the science rob paper, I think things got better when we articulated SPP in GCS vs GCSTrajOpt. Are we calling this GcsTrajOpt, too (e.g. GcsTrajOpt can now do planning through contact), or branding a new name? I know we're not doing bezier yet, but we could...}
%\bpgcomment{I don't fully understand this comment. Are asking whether we should use the Motion Planning formulation rather than the SPP in GCS formulation?}
The \acrfull*{spp} in \acrshort*{gcs} problem can then be stated as:
%\begin{subequations}
%\begin{align}
%    \min \quad& \sum_{e := (u,v) \in \mathcal{E}_p} l_e(x_u, x_v) \\
%    \text{s.t.} \quad&
%    p \in \mathcal{P}, \\
%    \quad & x_v \in \mathcal{X}_v, \quad \forall x_v \in p, \\
%    \quad & (x_u, x_v) \in \mathcal{X}_e, \quad \forall e := (x_u, x_v) \in \mathcal{E}_p
%\end{align}
%\label{eq:gcs}
%\end{subequations}
\begin{subequations}
\begin{align}
    \minimize \quad& \sum_{v \in p} l_v(x_v) \\
    \subto \quad&
    p \in \mathcal{P}, \\
    \quad & x_v \in \mathcal{X}_v, \quad \forall x_v \in p \\
    \quad & (x_u, x_v) \in \mathcal{X}_e, \quad \forall e := (x_u, x_v) \in \mathcal{E}_p
\end{align}
\label{eq:gcs}
\end{subequations}

Problem~\eqref{eq:gcs} can be solved exactly as a \acrfull*{micp}.
Importantly, the problem formulation has a very tight convex relaxation, and can in many instances be solved to global optimality by solving the convex relaxation and performing a cheap rounding step on the integer variables, as shown in~\cite{marcucci2023motionplanning}.
In this work, we treat the \acrshort*{gcs} framework as a modelling language that allows us to formulate and efficiently solve the problem presented in \eqref{eq:gcs}.
Additionally, the \acrshort*{gcs} framework naturally gives us an upper bound on the optimality gap to a solution; Let \(C_\text{relax} \leq C_{\text{opt}} \leq C_\text{round}\) be the costs of the relaxation, the \acrshort*{micp}, and the rounded solution, respectively. The optimality gap \(\delta_\text{opt}\) can then be overestimated as
\begin{align}
    \delta_\text{opt}  =
    \frac{C_\text{round} - C_\text{opt}}{C_\text{opt}}
    \leq
    \frac{C_\text{round} - C_\text{relax}}{C_\text{relax}}
    = \delta_\text{relax}
    \label{eq:optimality_bound}
\end{align}

Finally, we note that the original problem description in~\cite{marcucci2023shortest,marcucci2023motionplanning} has costs on edges rather than vertices, of which vertex costs is a special case.

\subsection{Semidefinite relaxation of quadratically constrained quadratic programs}
\label{sec:sdp_relaxation_of_qcqp}
%!TEX root = ../root.tex

%\tmcomment{This title is a little confusing: QCQPs can have affine constraints, so I would just say ``Semidefinite relaxation of quadratically constrained quadratic programs''}
%\tmcomment{It'd be good do have an introductory sentence with maybe a reference, an hugly version of which can be ``In this section we review QCQPs and their semidefinite relaxation, see~[reference] for a more detailed discussion.''}
In this section we review nonconvex \acrshort*{qcqps} and their semidefinite relaxation~\cite{shorQuadraticOptimizationProblems1987}, and refer the reader to~\cite{parkGeneralHeuristicsNonconvex2017,parriloLectureNotesMIT2023,sheraliReformulationLinearizationTechniqueSolving1999} for a more detailed discussion.
This semidefinite relaxation and related techniques are commonly used in combinatorial optimization (see, e.g.,~\cite{goemans1995improved}), and have been used in robotics for both localization and pose estimation~\cite{yangTEASERFastCertifiable2021,dumbgenSafeSmoothCertified2023,giamouSemidefiniteRelaxationsGeometric2023a}, as well as motion planning around obstacles~\cite{el2021piecewise}.

%In this section, we review the standard, nonconvex \acrshort*{qcqp} which we will use to model the control of our slider-pusher system through contact in \cref{sec:mechanics_of_planar_pushing}.
%As \acrshort*{qcqps} are known to be NP-hard to solve in general \cite{parkGeneralHeuristicsNonconvex2017}, we will also review a well-known \acrshort*{sdp}relaxation that will enable us to efficiently lower bound the cost and obtain good initial guesses for solving \acrshort*{qcqp} using a local optimizer;
%a more detailed discussion of \acrshort*{sdp} relaxations is available in \cite{parriloLectureNotesMIT2023, parkGeneralHeuristicsNonconvex2017,sheraliReformulationLinearizationTechniqueSolving1999}.
%
%\rtcomment{you're devoting a lot of space/attention to this if it's textbook.  i'm not sure we need it?  (unless we are clearly departing from the textbook, e.g. if/when we include support for SOCP constraints)? Maybe you don't need to cut it yet, but if we start space optimizing in the final stretch, I think it's a natural place for a haircut.}
%\tmcomment{I don't know why people call this ``homogeneous form'', $x_0=1$ is not homogeneous?} form as:
%\aacomment{Response to Tobia. This is called the "homogeneous form" in the sense that it is the "homogeneous coordinates". Essentially the original problem is the projective space of the homogeneous coordinates.}
Let $y \in \mathbb{R}^{n}$ and let $x \coloneqq \begin{bmatrix} 1 & y^{\intercal} \end{bmatrix}^{\intercal}$. A nonconvex \acrshort*{qcqp} in homogeneous form is the optimization program
\begin{subequations}
\begin{align}
    \minimize \quad x^\intercal Q_0 x & \\
    \subto
    \quad
    x^\intercal Q_i x &\geq 0, \quad \forall i = 1, \ldots, l \label{eq:qcqp_quad_constraint} \\
    \quad
    Ax &\geq 0 \label{eq:qcqp_affine_constraint} 
    % \\
    % \quad
    % x &=
    % \begin{bmatrix}
    % 1 \\
    % y
    % \end{bmatrix}
    % \label{eq:qcqp_x_constraint}
\end{align}
\label{eq:qcqp}
\end{subequations}
%\tmcomment{I would not specify the arguments in the optimization problem (under the min). But if you do, note that also x is a decision variable here. Also ``minimize'' is a slightly more correct than ``min''.}
\noindent where 
% \(x \in \mathbb{R}^{n+1}\), \(y \in \mathbb{R}^n\),
\(A \in \mathbb{R}^{m \times (n+1)}\), and \(Q_i \in \mathbb{R}^{(n+1) \times (n+1)}, \, i = 0, \ldots, l\). Note that positive-semidefiniteness is not required for \(Q_i\), hence \eqref{eq:qcqp} can be nonconvex.
% and in general NP-hard~\cite{parkGeneralHeuristicsNonconvex2017} to solve. \smcomment{NP hardness mentioned already in previous parargraph}
% Denote the feasible set of the \acrshort*{qcqp} in \eqref{eq:qcqp} by $\mathcal{S} := \set{ (x,y) \mid \eqref{eq:qcqp_quad_constraint} \ \text{to} \ \eqref{eq:qcqp_x_constraint}}$.
A lower bound to \eqref{eq:qcqp} can be obtained by solving the semidefinite program \cite{shorQuadraticOptimizationProblems1987, parriloLectureNotesMIT2023}:
\begin{subequations}
\begin{align}
    \minimize \quad
    \trace{(Q_0 X)} \\
    \subto
    \quad
    \trace{(Q_i X)} &\geq 0, \quad \forall i = 1, \ldots, l \label{eq:qcqp_relax_feas_1} \\
    AXe_1 &\geq 0 \label{eq:qcqp_relax_feas_2} \\
    AXA^\intercal &\geq 0 \label{eq:redundant_constraint} \\
    X &=
    \begin{bmatrix}
        1 & y^\intercal \\
        y & Y
    \end{bmatrix} \succeq 0
    \label{eq:big_X_in_SDP}
\end{align}
\label{eq:qcqp_relaxation}
\end{subequations}
where \(e_1\) is the \((n+1)\)-vector with a 1 as the first component, and all the rest being zero.

Recalling that if the PSD matrix $X$ is rank one, then $X = xx^{T}$, and using the simple identity \(x^\intercal Q_i x = \trace{(Q_i x x ^\intercal)}\), it can be seen that any feasible, rank-one solution to \eqref{eq:qcqp_relaxation} provides a feasible solution to \eqref{eq:qcqp}. Therefore \eqref{eq:qcqp_relaxation} is indeed a relaxation. Moreover, if $X^{*}$ is a rank-one minimizer of \eqref{eq:qcqp_relaxation}, then $x^{*}$ satisfying $X^{*} = x^{*}(x^{*})^\intercal$ is the global optimum of \eqref{eq:qcqp}.
This motivates the following scheme for obtaining near-optimal solutions to \eqref{eq:qcqp}:
After solving \eqref{eq:qcqp_relaxation} to obtain an $X^{*}$, we take $\hat{x} = \begin{bmatrix} 1 & \hat{y}^{\intercal}\end{bmatrix}^{\intercal}$ to be the first column of $X^{*}$.
% we compute the rank-one projection $\hat{x} = \begin{bmatrix} 1 & \hat{y}^{\intercal}\end{bmatrix}^{\intercal}$ such that $\hat{x} \approx X^{*}$ \aacomment{Bernhard do you use the maximum eigenvector or the last first column here?}\bpgcomment{First column}.\smcomment{that's less principled,no? do you get better performance with 1st column?}\smcomment{for paper sake might be easier to say you do rank 1 projection} 
We then use $\hat{x}$ as an initial guess for solving \eqref{eq:qcqp} using a local nonconvex solver. If a feasible solution is found, then we have the bound $C_\text{relax} \leq C_{\text{opt}} \leq C_\text{round}$ where $C_{\text{relax}}$, $C_{\text{opt}}$, and $C_\text{round}$ are the optimal costs of the semidefinite relaxation, the nonconvex \acrshort*{qcqp}, and the rounded solution, respectively. 
While this strategy is not guaranteed to find a feasible solution to \eqref{eq:qcqp}, our results in \Cref{sec:experiments} demonstrate that in the case of planar pushing, this strategy finds both feasible and near-optimal solutions for all the tested problem instances.

\section{The Slider-Pusher System}
\label{sec:mechanics_of_planar_pushing}
%!TEX root = ../root.tex

This section introduces the model for the slider-pusher system.
We first describe the kinematics for the slider-pusher system, and then describe the model of the contact interaction between the slider and the underlying surface, known as the limit surface. Then, we present the full quasi-static dynamics model along with the frictional constraints.
%\bpgcomment{The slider-pusher system for one contact can be shown to be differentiably flat \cite{lefebvreDifferentialFlatnessSliderPusher2023,zhouPushingRevisitedDifferential2019}. We do not leverage this. Importantly because it seems very problem specific, but also because it is probably not possible /easy to use with gcs due to flat coordinates not having a nice physical interpretation}

We use the monogram notation from~\cite[\S 3.1]{manipulation} for reference frames. Denote the position of the frame $B$ relative to $A$, measured in frame $C$ by $\prescript{A}{}{p}^B_C$ (when $C = A$, we write simply $\prescript{A}{}{p}^B$).
We similarly write \( \prescript{A}{}{v}^B_C\) (or \(\prescript{A}{}{v}^B\)) for translational velocities,
and
\(\prescript{A}{}{V}^B_C\)
(or \(\prescript{A}{}{V}^B\)) for spatial velocities.
The rotation that transforms a vector from frame \(B\) to frame \(A\) is denoted by \(\prescript{A}{}{R}^B\).
To denote a Cartesian force applied at a point $p$ measured in frame $A$ we will write $f^p_A$, and to denote the spatial force we write \(F^p_A\).
We will drop the frames when they are clear from the context.

%\subsection{Quasi-static assumption}
%Consider a dynamical system that is in contact with its environment at multiple contact points. Its dynamics are given by:
%\begin{align}
%M (q) \dot{{v}} + {C} (q, v) v = &\tau_g(q) + \sum_{i} J_{c,i}^\intercal f^{c_i}
%\end{align}
%where \(q \in \R^{n_q}\)
%and \(v \in \R^{n_v}\)
%are the generalized coordinates and velocities, 
%\(M, C \in \R^{{n_v}\times{n_v}}\) 
%are the mass and Coriolis
%matrices,
%\(\tau_g \in \R^{n_v}\) is the generalized gravity vector,
%and \(f^{c_i}\) and \(J_{c,i}\) are the contact force and contact Jacobian at contact \(i\). Under the quasi-static assumption, the velocities and accerelations in the system are low, implying that the inertial forces are dominated by frictional forces:
%\(M (q) \dot{{v}} + {C} (q, v) v \approx 0\)
%, which gives the quasi-static dynamics:
%\begin{align} 
%\tau_g(q) + \sum_i J_{c,i}^\intercal f^{c_i} = 0
%\end{align}
%\bpgcomment{Do we really need this paragraph?}

\subsection{Kinematics}
\begin{figure}[ht]
    \begin{subfigure}[b]{0.64\linewidth}
    	\centering
    	\includegraphics[width=\linewidth]{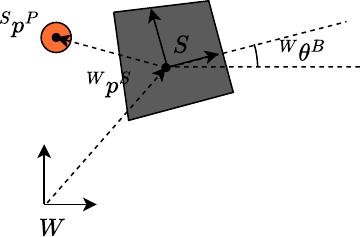}
        \caption{}
    	\label{fig:planar_pushing_kinematics}
    \end{subfigure}
    \centering
    \begin{subfigure}[b]{0.27\linewidth}
        \centering
        \raisebox{49pt}{
        \includegraphics[width=\linewidth]{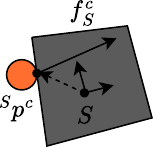}
        }
        \caption{}
        \label{fig:slider_pusher_forces}
    \end{subfigure}
    \caption{a) The slider-pusher kinematic quantities. b) The contact point and the contact forces.}
\end{figure}

Consider the slider-pusher system in \Cref{fig:planar_pushing_kinematics}. Let \(W\) be the world frame, and attach the reference frames \(S\) and \(P\) at the \acrshort*{com} of the slider and pusher, respectively. Let the pose of the slider in the world frame be given by
\((\prescript{W}{}{p}^S, \prescript{W}{}{\theta}^S) \in \mathrm{SE}(2)\).
%\rtcomment{can we please use the formatting e.g. $\mathrm{SE}(2)$ throughout?}
Let the (planar) spatial velocity of the slider be
\(\prescript{W}{}{V}^S = 
(
\prescript{W}{}{v}^S,
\prescript{W}{}{\omega}^S
) \in \mathrm{se}(2)\) (the tangent space associated with $\mathrm{SE}(2)$), where
\( \prescript{W}{}{v}^S = (d/dt) \prescript{W}{}{p}^S \) and 
\(\prescript{W}{}{\omega}^S = 
(d/dt) \prescript{W}{}{\theta}^S\).
Similarly, we represent the position of the $i$-th vertex of the slider with $\prescript{W}{}{p}^{\nu_i}$, and its velocity by $\prescript{W}{}{v}^{\nu_i}$, where $i = 1, \ldots, N_\nu$, and $N_\nu$ denotes the number of slider vertices.
We represent the position of the pusher in the frame of the slider by 
\(\prescript{S}{}{p}^P \in \R^2\), and denote its velocity by \( \prescript{S}{}{v}^P = (d/dt) \prescript{S}{}{p}^P\).
We do not assign an orientation to the pusher, as it is assumed to be circular and hence invariant to rotation. 
When there is contact between the slider and the pusher, we will denote the position of the contact point in the slider frame by \(\prescript{S}{}{p}^c \in \R^2\) and the velocity by $\prescript{S}{}{v}^c = (d/dt) \prescript{S}{}{p}^c$
%\rtcomment{this is true in the plane, but not true with quaternions, of course}.
%\bpgcomment{Did you think I should change something here?}
Going forward we will use these definitions, but will often drop the frames to keep notation light.
We use \((p^S, \theta)\) and \( V = (v^S, \omega^S)\) to denote the slider pose and spatial velocity in the world frame,
and \(p^P\), \(v^P\), \(p^c\), and $v^c$ to denote the pusher position, translational pusher velocity, contact point position, and contact point velocity in the frame of the slider.
%\rtcomment{in which frame? i know the notation is heavy, but this seems like a slippery slope. If leaving the frames off always implies world frame, then that's consistent with convention and is not simply dropping the frames for simplicity.}.
%\bpgcomment{Better now?}

%!TEX root = ../root.tex
%\begin{figure}[ht]
%    \begin{subfigure}[b]{0.45\linewidth}
%    	\centering
%    	%\includegraphics[width=\linewidth]{media/planar_pushing_kinematics.pdf}
%            \input{figures/tikz_figures/pusher_slider/pusher_slider_kinematic_quantities}
%        \caption{}
%    	\label{fig:planar_pushing_kinematics}
%    \end{subfigure}
%    %
%    \hfill
%    \centering
%    \begin{subfigure}[b]{0.45\linewidth}
%        \centering
%        % \includegraphics[width=\linewidth]{media/planar_pushing_kinematics.pdf}
%        \input{figures/tikz_figures/pusher_slider/pusher_slider_forces}
%        \caption{\bpgcomment{TODO finalize these figures}\aacomment{Can you send me a sketch?}}
%        \label{fig:slider_pusher_forces}
%    \end{subfigure}
%    \caption{a) The slider-pusher kinematic quantities. b) The contact point and the contact forces.}
%\end{figure}
%
%
%

%!TEX root = ../root.tex
%In the final version, these figures should be bigger.

\subsection{The limit surface}
The limit surface is a convex geometric object that describes the relationship between the applied spatial contact force on an object that is resting on a frictional surface and its instantaneous spatial velocity under the assumption of quasi-static dynamics~\cite{goyalPlanarSlidingDry1991}.
In this work, we adapt the commonly used ellipsoidal approximation of the limit surface
to model the interaction between the contact force applied by the pusher and the resulting spatial slider velocity~\cite{ hoganReactivePlanarNonprehensile2020, lynchManipulationActiveSensing1992, chavan-daflePlanarInhandManipulation2020}.

Let \(f^c_S \in \R^2\) be the contact force applied by the pusher measured in the slider frame, and let \( \tau^c_S \in \R\) be the resulting contact torque. 
Let \(F^c_S \in \R^3\) denote the (planar) spatial contact force, given by:
\begin{align}
    F^c_S =
    \begin{bmatrix}
       f \\
       \tau
    \end{bmatrix}
    = (J^c_S)^T f,
    \quad
    J^c_S =
    \begin{bmatrix}
        1 & 0 & -p^c_y \\
        0 & 1 & p^c_x \\
    \end{bmatrix}
    \label{eq:spatial_contact_force}
\end{align}
where \(J^c_S \in \R^{2\times 3}\) denotes the contact Jacobian, and the frames have been dropped on the contact force and torque.
%\aacomment{When defining $J_{c}$ and $F_{S}^{c}$ I would not drop the frames yet. Also I think its more correct to use $J^{c}$ even if its annoying to write $(J^{c})^{T}$.}
From now, we will write simply \(f\), \(\tau\), and \(F\). See~\Cref{fig:slider_pusher_forces} for an illustration of these quantities.

We let the \textit{limit surface} be defined by the ellipsoid in \(\R^3\) described by \( H(F) = (1/2) F^T D F = 1\), where
\begin{align}
    D &= \text{diag}(c_f^{-1}, c_f^{-1}, c_\tau^{-1}) \in \R^{3\times3} \\
    c_f &= \mu_S m_S g, \, \,
    c_\tau = c r \mu_S m_S g
    \label{eq:limit_surface_const}
\end{align}
and \(\mu_S \in \R\) is the friction coefficient between the slider and table, \(m_S\in\R\) is the
slider mass, $g \in \R$ is the gravitional acceleration,
\(r \in \R\) is a characteristic distance, typically chosen as the max distance between a contact point and origin of frame $S$,
and \( c \in [0,1]\) is an integration constant that depends on the slider geometry, as detailed in~\cite{lynchManipulationActiveSensing1992} and~\cite{chavandafleInHandManipulationMotion2018}.
From the principle of maximum dissipation, it can be shown
that when the slider is sliding, the applied spatial contact force \(F\) must lie on the limit surface, and the instantaneous spatial velocity \(V\) must be perpendicular to the limit surface:
\( V \propto \nabla H(F) = DF \)
, where \( \nabla (\cdot)\) denotes the gradient operator and \(F\) and \(V\) are measured in the same frame~\cite{goyalPlanarSlidingDry1991}.
We note that this can be approximated as the linear constraint
\( V = DF \) and appropriately scaling \( F \) as a post-processing step.
%\rtcomment{is that true? is this the only place F enters? what about the objective terms using F?}.
%\bpgcomment{That is a good point. The regularization on the contact forces in the cost term will therefore be coupled with the slider velocity through $V = DF$.}
%\bpgcomment{Is it confusing that I use the propto symbol here (could be interpreted as the velocity growing with the gradient of the limit surface, which doesn't make much sense)? What would a better symbol be?}
%\aacomment{We could write $V \in \textbf{span}H(F)$ or $V \perp H(F)$. For example "From the principle of..." $V$ be orthogonal to the limit surface (i.e.) $V \perp H(F)$. With appropriate scaling of $F$, this can be enforced as the linear equality constraint $V = DF$.}

\subsection{Slider-pusher dynamics}
Define the state of the slider-pusher system as \(x = (p^S, \, \theta, \, p^P)\) and the input as \(u = (f, v^P)\). 
Let \( \phi : \R^2 \rightarrow \R\)
denote the \acrfull*{sdf} between the slider and pusher, a function of only the pusher position in the slider frame.
Using the limit surface approximation from the previous subsection, the dynamics for the slider-pusher system can be written as:
\begin{align}
    \dot x &= g(x,u)
    =
    \begin{bmatrix}
    R V_S \\
    v^P
    \end{bmatrix}
    =
    \begin{bmatrix}
    R D F_S
    \label{eq:slider_pusher_dynamics}
    \\
    v^P
    \end{bmatrix}
    \\
    0 &\leq \phi(p^P) \perp \| f \|_2 \geq 0
    \label{eq:contact_compl_constraint}
\end{align}
%\tmcomment{I would have immagined that this part reconnected to subsection VIA (the one with the quasistatic dynamics)? If that subsection is not used anywhere, we can cut it as per your comment above?}
%\bpgcomment{Oops, there is a notation clash with \(f\) for force and dynamics. How about using i) $g$ for dynamics, $f$ for force, and $F$ for spatial force, or ii) $\lambda$ for forces, $\Lambda$ for spatial forces, and $f$ for dynamics?}
where
$$
R = 
\begin{bmatrix}
    \prescript{W}{}{R}^S & 0 \\ 
    0 & 1 \\
\end{bmatrix}, \quad
\prescript{W}{}{R}^S \in \mathrm{SO(2)}
$$
transforms the spatial velocity from the slider frame into the world frame. Condition~\eqref{eq:contact_compl_constraint} is a complimentarity constraint that captures the fact that there can be no contact force when the pusher is not in contact with the slider.

\subsection{Frictional constraints}
The dynamics in~\eqref{eq:slider_pusher_dynamics} and~\eqref{eq:contact_compl_constraint} do not enforce that the contact force between the slider and pusher lies in the friction cone.
Let \( \lambda_n, \lambda_f \in \R\) be the normal and frictional components of \(f\), that is, the components perpendicular and parallel to the contact face, respectively. Coloumb's friction law then states:
\begin{align}
    \lambda_n \geq 0, \quad |\lambda_f| \leq \mu \lambda_n
    \label{eq:friction_cone}
\end{align}
where \(\mu \in \R\) is the friction coefficient between the slider and pusher.
Note that, subject to~\eqref{eq:friction_cone}, the condition \(\|f\|_2 = 0\) from the complimentarity constraint in~\eqref{eq:contact_compl_constraint} is equivalent to \(\lambda_n = 0\).

Further, Coulomb's friction law states that when there is sliding motion between two objects, the friction force must be opposing the direction of motion, and lie on the boundary of the friction cone. Let $\prescript{S}{}{v}^{c_\perp} \in \R$ be the tangential contact velocity in the slider frame. The contact modes are then given by:
\begin{itemize}
\item 
\emph{Sticking.} When the contact between the pusher and slider is sticking, the relative tangential velocity must be zero
\begin{align}
\prescript{S}{}{v}^{c_\perp} = 0
\label{eq:mode_sticking}
\end{align}
\item 
\emph{Sliding Left.} When the contact point is sliding left relative to the slider, the friction force must lie on the boundary of the friction cone:
\begin{align}
\prescript{S}{}{v}^{c_\perp}
\leq 0, \quad
\lambda_{f} = -\mu \lambda_{n}
\label{eq:mode_sliding_left}
\end{align}
\item
\emph{Sliding Right.} When the contact point is sliding right, the relationship is flipped:
\begin{align}
\prescript{S}{}{v}^{c_\perp}
\geq 0, \quad
\lambda_{f} = \mu \lambda_{n}
\label{eq:mode_sliding_right}
\end{align}
\end{itemize}

\section{Motion Planning for Planar Pushing}
\label{sec:motion_planning}
%!TEX root = ../root.tex
In this section, we present the motion planning algorithm.
First, we decompose the configuration space of the slider-pusher system into a collection of subsets that we call \textit{modes}.
For each face of the slider, we have two types of modes: contact and non-contact.
We then show how to formulate the motion planning problem within a mode as a convex program.
Next, we formulate the motion planning problem as an \acrshort*{spp} in a \acrshort*{gcs}, 
by constructing a graph where vertices correspond to modes of the system, edges correspond to valid mode transitions, and a path corresponds to a mode sequence and the motion for each mode in the sequence.

\subsection{Modes of the slider-pusher system}
\label{sec:modes}
%\tmcomment{Similar to above: the term ``mode'' is informal, and does not really has a mathematical meaning. It should be clearly defined--probably as a region that belongs to some sort of decomposition of the space.}
%!TEX root = ../root.tex

\begin{figure}
    \begin{subfigure}[b]{0.50\linewidth}
        \centering
        \includegraphics[width=\linewidth]{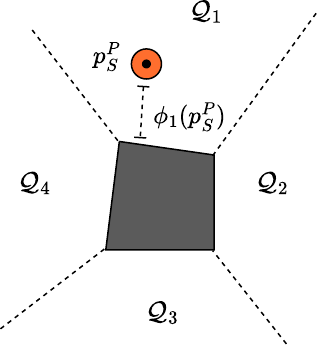}
        \caption{}
        \label{fig:configuration_space}
    \end{subfigure}
    \begin{subfigure}[b]{0.40\linewidth}
        \centering
        \includegraphics[width=\linewidth]{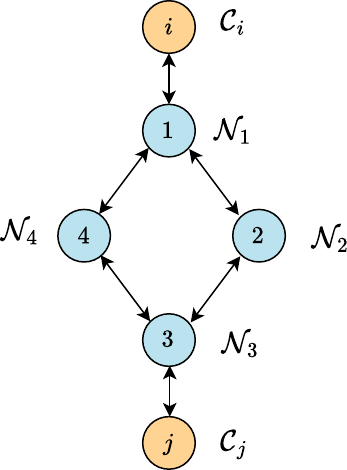}
        \caption{}
        \label{fig:subgraph}
    \end{subfigure}
    \caption{a) An example of a configuration-space partitioning \(\mathcal{Q}_1, \ldots, \mathcal{Q}_4\) and the linear approximations \(\phi_1, \ldots, \phi_4\) for a slider with convex planar geometry. b) The graph of non-contact modes that is added between every pair of vertices corresponding to contact modes $\mathcal{C}_i$ and $\mathcal{C}_j$. Corresponding modes are written in text next to the vertices.
    }
    
\end{figure}

The complimentarity constraint in~\eqref{eq:contact_compl_constraint} describes two classes of modes for the system: \textit{non-contact}
%\tmcomment{Collision or contact? I think the latter is more accurate?}
when \(\phi(p^P) > 0\) and \textit{contact} when \(\phi(p^P) = 0\).  
The planar slider geometry is assumed a polygon, and we denote its $N_F$ faces by 
\(\mathcal{F}_1, \ldots, \mathcal{F}_{N_F}\).
We can then partition the collision-free configuration space for the pusher \(\mathcal{Q} = \set{p^P \mid \phi(p^P) \geq 0} \subseteq \R^2\) into \( N_F\) collision-free, possibly overlapping, polyhedral regions \(\mathcal{Q}_1, \ldots, \mathcal{Q}_{N_F} \subseteq \mathcal{Q}\)
in such a way that for each $\mathcal{Q}_i, \ i = 1, \ldots, N_F$ one face of the slider is represented by a supporting hyperplane to $\mathcal{Q}_i$.
%in such a way that each region intersects at least one face of the slider, i.e.,
%\(\mathcal{F}_i \subseteq \mathcal{Q}_j\)  for all \(i=1, \ldots, N_F\) and some \(j \in \set{1, \ldots, N_F}\).
%\aacomment{Intersects the face might give people the impression that regions can penetrate into the slider. The convex geometry way of saying this would be, "we assume that for each region $\mathcal{Q}_{i}$, at least one face of the slider is a supporting hyperplane to $\mathcal{Q}_{i}$"}
We additionally approximate $\phi$ by a piecewise linear function according to the decomposition \(\mathcal{Q}_1, \ldots, \mathcal{Q}_{N_F}\) 
by associating a linear function \(\phi_i\) to each \(\mathcal{Q}_i, \, i = 1, \ldots, N_F\).

In~\Cref{fig:configuration_space} we provide an example for a convex slider geometry,
where 
one collision-free region is defined as the outside of each face such that two adjacent sets \(\mathcal{Q}_i\) and \(\mathcal{Q}_j\) are ``split" by the vector \((\hat{n}_i + \hat{n}_j)/2\)
%\aacomment{Doesn't this depend on where you attach the normal vectors to the face?}
%\bpgcomment{Vectors don't have a position, only a direction and magnitude, so I don't think so}
, where \(\hat{n}_i, \, \hat{n}_j\) denotes the normal vectors of face \(\mathcal{F}_i\) and \(\mathcal{F}_j\). 
We can then define \(\phi_i\) as the Euclidean distance from \(p^P\) to the half-plane that describes \(\mathcal{F}_i\).
When the closest point on the slider to \(p^P\) is on a face we have \(\phi_i(p^P) = \phi(p^P)\), and \(\phi_i(p^P) = 0 \) only when the pusher is in contact with the slider.
We note that in the case of a non-convex slider geometry a similar (but slightly more complex) decomposition of \(\mathcal{Q}\) and approximation of \(\phi\) is possible.
%\aacomment{To save space, it may be okay to punt this discussion purely to a picture.}
%\bpgcomment{What do you think? Cut this/move to a picture/keep it?}
%\aacomment{Yes I would be the final sentence here to the caption. of figure 3}

When the slider is not in contact with the pusher, the contact force must be zero due to~\eqref{eq:contact_compl_constraint}, and~\eqref{eq:slider_pusher_dynamics} then implies that the slider pose is constant.
For each of the collision-free configuration-space regions \(\mathcal{Q}_i\) we therefore define a non-contact mode \(\mathcal{N}_i\) by the constraint:
\begin{subequations}
\begin{align}
    p^P &\in \mathcal{Q}_i
    \label{eq:collision_free_region_constraint}
\end{align}
\label{eq:non_contact_constraint}
\end{subequations}
and the last row of~\cref{eq:slider_pusher_dynamics}.
Hence, for the non-contact modes
we do not model the slider velocity or contact forces, and 
given a feasible initial slider pose that satisfies~\eqref{eq:so2}, the non-contact dynamics are described by linear constraints.

Similarly, for each face \(\mathcal{F}_i, \, i = 1, \ldots, N_F\), on the planar slider geometry we define a contact mode \(\mathcal{C}_i\) by the constraints \eqref{eq:spatial_contact_force}, \eqref{eq:slider_pusher_dynamics},~\eqref{eq:friction_cone},~\eqref{eq:mode_sticking}, and
\begin{align}
    \phi_i(p^P) = 0
    \label{eq:contact_constraints}
\end{align}
For simplicity, we do not consider sliding contact modes between the pusher and slider.

Finally, we represent \(\theta^S\)
%\aacomment{I don't think you have said that you are parametrizing rotations with $\theta$ yet}
%\bpgcomment{Yes its in the kinematics section}.
with two variables \(c_\theta, s_\theta \in \R\), with the additional constraint:
\begin{align}
    % \| r \|_2^2 = 1, 
    r \coloneqq (c_\theta, s_\theta),
    \quad
    \| r \|_2^2 = 1
    % r := (c_\theta, s_\theta)
    \label{eq:so2}
    \iff
    \begin{bmatrix}
        c_\theta & -s_\theta \\
        s_\theta & c_\theta \\
    \end{bmatrix}
    \in \mathrm{SO(2)}
\end{align}
%With this representation, the angular velocity of the slider can be expressed in terms of $c_\theta$ and $s_\theta$ as $\omega^S = c_\theta \dot{s_\theta} + s_\theta \dot{c_\theta}$.
It now follows that the non-contact modes are described by linear constraints, and the contact modes are described by linear and (non-convex) quadratic constraints.

\subsection{Motion planning within a mode}
%We seek a cost on the total trajectory arc length \(\int^T_0 \| \dot x(t) \|_2 dt\), and a cost on the trajectory energy \( \int^T_0 \| \dot x(t) \|_2^2 dt \) to regularize the input and velocities of the system.
%\tmcomment{I don't fully understand this part. Is this the objective function? If yes, why then you give two different objectives for contact and no contact? Can one skip this?}
%Here, \(T \in \R_{>0}\) denotes the trajectory duration.
We discretize the dynamics in~\eqref{eq:slider_pusher_dynamics} using forward Euler discretization. We represent a trajectory segment within each mode for the slider-pusher system by $N$ discrete knot points for the state and $N-1$ knot points for the input: \(x_0, x_1, \ldots, x_N\) and \(u_0, u_1, \ldots, u_{N-1}\). The dynamics are then \(x_{k+1} = x_k + h g(x_k, u_k)\) for \(k = 0, \ldots, N-1\) with the chosen time step \(h \in \R_{>0}\).

Our formulation allows any cost that is a combination (or subset) of arc length and trajectory energy for the slider and pusher, contact forces, and time spent in (and near) contact.
The cost takes the following form:
\begin{align}
    \sum_{k=0}^{N-1}
    &L(x_k, x_{k+1})
    +
    E(x_k, x_{k+1}, u_k) \nonumber \\
    &+ 
    k_{f} h \| f_k \| _2^2
    + \sum_{k=0}^{N} 
    \psi(x_k)
    \label{eq:cost}
\end{align}
where the terms model the total trajectory arc length on $\mathrm{SE}(2)$ for the pusher and slider, the total kinetic energies for the pusher and slider trajectories, contact force regularization, and time spent near contact, respectively. Specifically
\begin{align}
    L(x_k, x_{k+1}) &=  k_{p^P} \| p^P_{k+1} - p^P_k \|_2 \nonumber \\
    &+ k_{p^S} (1/N_\nu) \sum_{i=1}^{N_\nu} \| p_{k+1}^{\nu_i} - p_k^{\nu_i} \|_2
\end{align}
models the total arc lengths on $\mathrm{SE}(2)$ of the pusher and slider, and 
\begin{align}
    E(x_k, x_{k+1}, u_k) &= k_{v^P} \| v^P_{k} \|_2^2 \nonumber \\
    &+ k_{v^S} (1/N_\nu) \sum_{i=1}^{N_\nu} \| v_{k}^{\nu_i} \|_2^2 
\end{align}
models the total kinetic energies of the trajectories for the pusher and slider. 
Recall that $p^{\nu_i}$ and $v^{\nu_i}$ refer to the position and velocity for slider vertex $i = 1, \ldots, N_\nu$ in the world frame, where $N_\nu$ is the number of slider vertices. The total arc length of the slider trajectory on $\mathrm{SE}(2)$ is modelled through the mean arc length traversed by its vertices. Likewise, the energy of the slider trajectory is modelled through the mean squared velocity of its vertices.
The constants \(k_{p^P}, \, k_{p^S}, \, k_{v^S}, \, k_{v^P}\) and \(k_f \in \R_{>0}\) are cost weights for the arc lengths for the pusher and slider, trajectory energies for the pusher and slider, and contact force regularization.

The last term in~\eqref{eq:cost} is a penalty on the distance between the pusher and the closest face of the slider.
It is given by:
\begin{align}
    \psi(x) = \max_{i=1, \ldots, N_F} \frac{h k_T}{1 + \frac{1}{k_\phi} \phi_i(p^P)}
\end{align}
where $k_T, \, k_\phi \in \R_{>0}$ are cost weights for the time in contact and for being ``close" to the slider, respectively.
This term is at its maximum when there is contact between the pusher and slider, and increases monotonically as the pusher gets closer to the slider.
It has the property that $\psi(x) = h k_T$ when $\min_i \phi_i(x) = 0$, i.e. the cost of contact between the pusher and slider is $k_T$ per second spent in contact. It also has the property that increasing $k_\phi$ increases the cost for being at a given distance to the object, i.e. $k_\phi$ controls the notion of ``close".
Since the denominator is always positive, this function is a maximum over convex functions and is thus readily encoded through $N_F$ \acrfull*{rsoc} constraints \cite{boydConvexOptimization}.

We define the motion planning problem for a contact mode as minimizing~\eqref{eq:cost} subject to~\eqref{eq:spatial_contact_force}, \eqref{eq:slider_pusher_dynamics},~\eqref{eq:friction_cone},~\eqref{eq:mode_sticking},~\eqref{eq:contact_constraints},~\eqref{eq:so2}, and initial and final conditions on the state.
This motion planning problem is non-convex due to the quadratic equality constraints arising from~\eqref{eq:spatial_contact_force},~\eqref{eq:slider_pusher_dynamics} and~\eqref{eq:so2}.
Specifically, the problem is a non-convex \acrshort*{qcqp}, which we relax into an \acrshort*{sdp} according to \Cref{sec:sdp_relaxation_of_qcqp}.
Further, we define the motion planning problem for a non-contact mode as minimizing~\eqref{eq:cost} subject
to the constraint~\eqref{eq:collision_free_region_constraint}, the last row of~\eqref{eq:slider_pusher_dynamics}, a constant slider pose, and initial and final conditions on the pusher position.
Since these are all linear constraints, motion planning for a non-contact mode is naturally a convex program.

\subsection{Constructing the graph of convex sets}
Finally, we formulate the motion planning problem as an \acrshort*{spp} in a \acrshort*{gcs}.
Let $G = (\mathcal{V}, \mathcal{E})$ be the graph underlying a \acrshort*{gcs} as defined in \Cref{sec:gcs}.
%\tmcomment{I would not say that $G$ is the GCS. I'd rather say that $G$ is the graph underlying the GCS, or domething like that.} \aacomment{Ditto with Tobia, this depends on how you define GCS. I would also point to the formal definition not a section.}
For every contact mode \(\mathcal{C}_i, \, i = 1,\ldots, N_F\) in the slider-pusher system, we add its corresponding \acrshort*{sdp} as a vertex \(v\) to \(\mathcal{V}\), in the sense that we let \(\mathcal{X}_v\) and \(l_v\) be the feasible set and cost function of the \acrshort*{sdp}, respectively. The point \(x_v \in \mathcal{X}_v\) now corresponds to a trajectory of length $N$ of states and inputs for the slider-pusher system in mode \(\mathcal{C}_i\).

Transitions between contact modes require the pusher to move within the non-contact modes, from one face of the slider to another. Since the pusher can visit every non-contact mode in each of these trajectories, we add to the graph a copy of all the non-contact modes for every possible contact-mode transition. Specifically, for any two contact modes $\mathcal{C}_i$ and $\mathcal{C}_j$ where $i, j \in \set{1, \ldots, N_F}$, $i \neq j$, we create a graph $G_{ij} = (\mathcal{V}_{ij}, \mathcal{E}_{ij})$
where $\mathcal{V}_{ij}$ contains a vertex $v$ for each non-contact mode $\mathcal{N}_k, \, k = 1, \ldots, N_F$ with  
$\mathcal{X}_v$ and $l_v$ taken as the feasible set and cost function for the mode, and 
$\mathcal{E}_{ij}$ contains a bi-directional edge between all non-contact modes with intersecting collision-free regions.
The vertices and edges of $\mathcal{G}_{ij}$ are added to $G$, as well as a bi-directional edge connecting the vertices corresponding to contact modes $\mathcal{C}_i$ and $\mathcal{C}_j$ to their respective non-contact modes to encode transition between contact and non-contact. 
An illustration is shown in~\Cref{fig:subgraph}.
The total number of vertices in the graph is then $\mathcal{O}((N_F)^3)$, with $N_F$ contact modes and $N_F \binom{N_F}{2}$ non-contact modes.

We enforce continuity between the state trajectories on a path in the graph. Specifically, for each edge \(e = (u,v) \in \mathcal{E}\)
we enforce that the last state in the trajectory in vertex \(u\) is equal to the first state in the trajectory in vertex \(v\).
Finally, we let the source vertex $s$ and target vertex $t$ be singleton sets that denote the desired starting and target states of the slider-pusher system, respectively.
A feasible path $p$ through $G$ then has the interpretation as a continuous trajectory from the initial state to the target state, that consists of distinct trajectory segments for each mode represented by the vertices on the path, and the trajectory within each mode is determined by the points in the vertices.

\subsection{Additional remarks}
\label{sec:tightnening_constraints}
In the semidefinite relaxation of the QCQP, \eqref{eq:redundant_constraint} are new constraints formed by multiplying the linear constraints from the original problem together. This is a general recipe; we can potentially continue to multiply constraints together to obtain tighter relaxations at the cost of generating higher-degree polynomial constraints, using the \acrfull*{sos} / moment hierarchy~\cite{lasserre2018moment}. Indeed, recent applications of the SOS hierarchy in motion planning have suggested that these higher-degrees were necessary to get tight relaxations~\cite{teng2023convex}. In contrast, in this work we are able to obtain tight relaxations using only quadratic constraints, avoiding the significant computational cost of using higher-order relaxations. We accomplish this by writing additional tightening constraints implied by~\cref{eq:so2}.
%Here, we add some selected higher-order tightening constraints, which enables us to obtain empirically tighter solutions without having to resort to a higher-order relaxation.
%We note that adding constraints like this is a procedure that we are currently working on automating.
%These are constraints that would be redundant in the original \acrshort*{qcqp}, but are not redundant for the relaxation, and hence potentially yields a smaller feasible set and a tighter relaxation.

First, we leverage~\eqref{eq:so2} (which, for general rotation matrices looks like $R^\intercal R = I$) and add the dynamic constraints in~\eqref{eq:slider_pusher_dynamics}
in both the world frame and the slider frame.
While these constraints are obviously redundant for the \acrshort*{qcqp} due to~\eqref{eq:so2}, they yield different constraints in the semidefinite relaxation.
Second, we use the initial and target rotations to generate tightening constraints.
In~\eqref{eq:so2} we parametrize the slider rotation by two variables in $\R^2$, where the start and target constraints on $\theta^S$ correspond to two points $r_s$ and $r_t$ on the unit circle.
We define a hyperplane between these points in $\R^2$ with the normal vector pointing away from the origin, and add the constraint that $r_k$ must be on the outside of this hyperplane for $k = 1, \ldots, N$.
Under the assumption that an optimal plan should always pick the shortest geodesic path between two elements of $\mathrm{SO(2)}$, this constraint is redundant for the \acrshort*{qcqp} where~\eqref{eq:so2} holds, but is not redundant in the relaxation.

Additionally, we exploit structure in our problem that allows us to write a computationally cheaper semidefinite relaxation in~\eqref{eq:qcqp_relaxation}.
Specifically, the motion planning problem for a contact mode exhibits a band-sparse structure, in the sense that the nonconvex quadratic constraints~\eqref{eq:spatial_contact_force},~\eqref{eq:slider_pusher_dynamics},~\eqref{eq:so2} only couple succeeding knot points $x_k$, $x_{k+1}$, and $u_k$, for $k = 0, \ldots, N-1$.
This allows us to form the relaxation~\eqref{eq:qcqp_relaxation} as an~\acrshort*{sdp} with several smaller~\acrshort*{psd} matrices instead of a single, larger matrix.
In principle, this does not include all the tightening constraints~\eqref{eq:redundant_constraint} and yields a potentially weaker convex relaxation, but in practice, we find that the loss in tightness is negligible while significantly reducing the computational cost.
For more details, the reader is referred to e.g. ~\cite{garstkaCliqueGraphBased2020}.

\section{Experiments}
\label{sec:experiments}
%!TEX root = ../root.tex

%!TEX root = ../root.tex
%\begin{figure}[t]
%	\centering
%    \begin{subfigure}[b]{\linewidth}
%    		\centering
%    		\includegraphics[width=\linewidth]{media/trajectory_box.pdf}
%       \caption{}
%      \end{subfigure}
%	\begin{subfigure}[b]{\linewidth}
%			\centering
%			\includegraphics[width=\linewidth]{media/trajectory_tee.pdf}
%       \caption{}
%    \end{subfigure}
%    \caption{Examples of motion plans for two differently shaped sliders.}
%\end{figure}
%\begin{figure*}[t]
%        \centering
%        %\includegraphics[width=1.0\linewidth]{media/run_18_trajectory.pdf}
%        \includegraphics[width=1.0\linewidth]{media/new_traj_figure.pdf}
%    \caption{Our planner simultaneously reasons about both discrete mode switches and continuous motion. Here, an example of a planar pushing plan with multiple mode switches is shown for a rectangular slider geometry.}
%    \label{fig:plan_example}
%\end{figure*}

\begin{figure*}[t]
        \centering
        \includegraphics[width=1.0\linewidth]{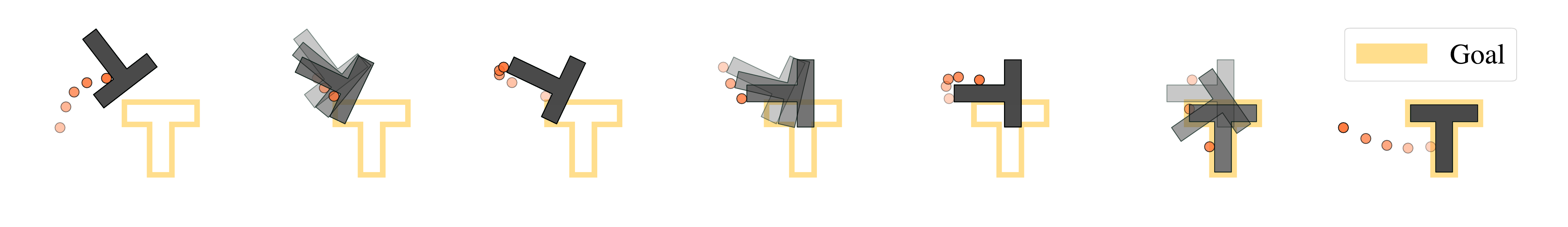}
    \caption{Our planner simultaneously reasons about both discrete mode switches and continuous motion. Here, an example of a planar pushing plan with multiple mode switches is shown for a T-shaped slider geometry.}
    \label{fig:plan_example}
\end{figure*}
%\input{figures/ablation_study}
%!TEX root = ../root.tex
\begin{figure*}[t]
	\centering
	\begin{subfigure}[b]{1.0\linewidth}
			\centering
			\includegraphics[trim=0 50 0 0, clip, width=\linewidth]{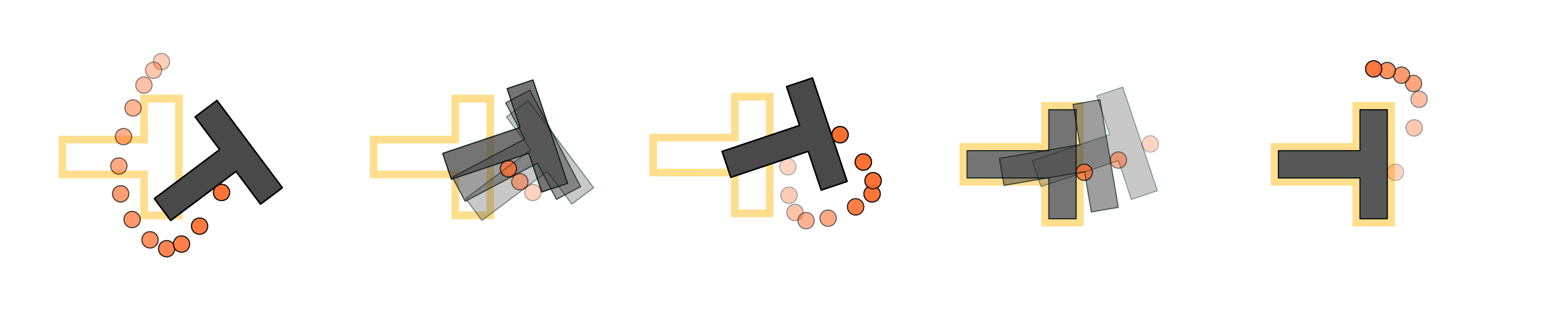}
    \end{subfigure}
	\begin{subfigure}[b]{1.0\linewidth}
			\centering
			\includegraphics[trim=0 0 0 20, clip, width=\linewidth]{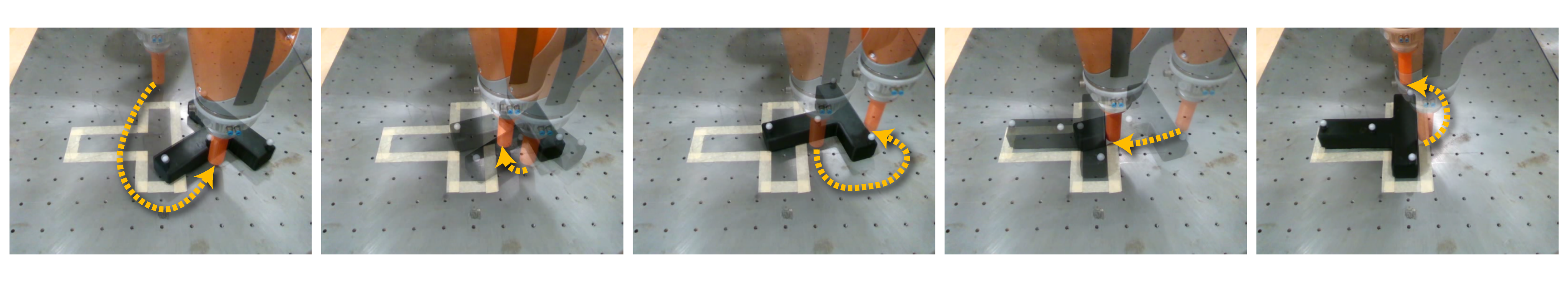}
    \end{subfigure}
	\begin{subfigure}[b]{1.0\linewidth}
			\centering
			\includegraphics[trim=0 50 0 0, clip, width=\linewidth]{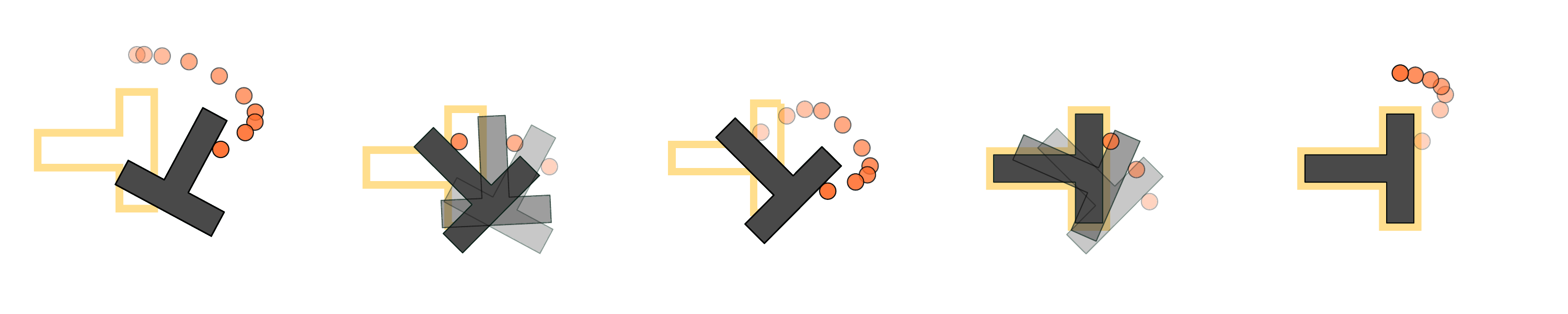}
    \end{subfigure}
	\begin{subfigure}[b]{1.0\linewidth}
			\centering
			\includegraphics[trim=0 0 0 20, clip, width=\linewidth]{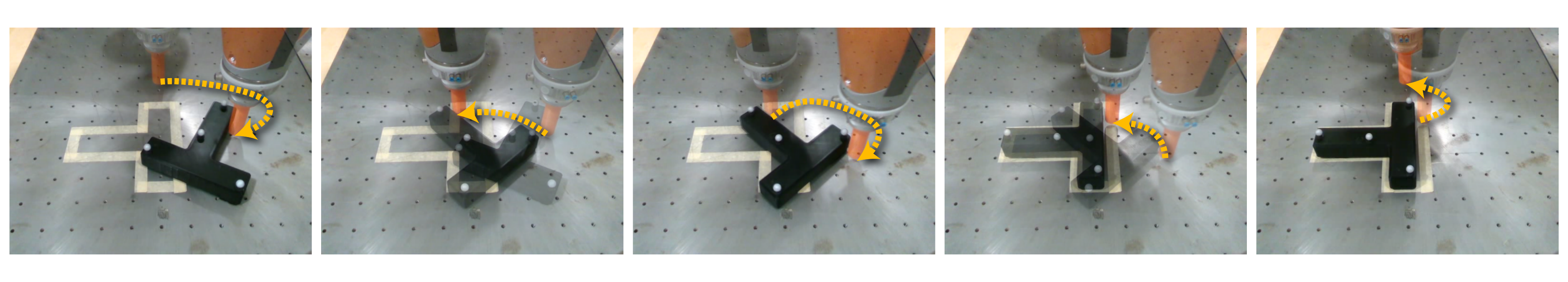}
    \end{subfigure}

   \caption{Our method is able to generate close-to globally optimal plans for pushing tasks with collision-free motion planning between contact modes. Here, two different pushing trajectories for a T-shaped slider are shown, stabilized with a feedback controller on a real robotic system.}
   \label{fig:experiments_and_plans}
\end{figure*}
%!TEX root = ../root.tex
\begin{figure*}[t]
	\centering
	\begin{subfigure}[b]{0.8\linewidth}
			\centering
			\includegraphics[trim=0 50 0 0, clip, width=\linewidth]{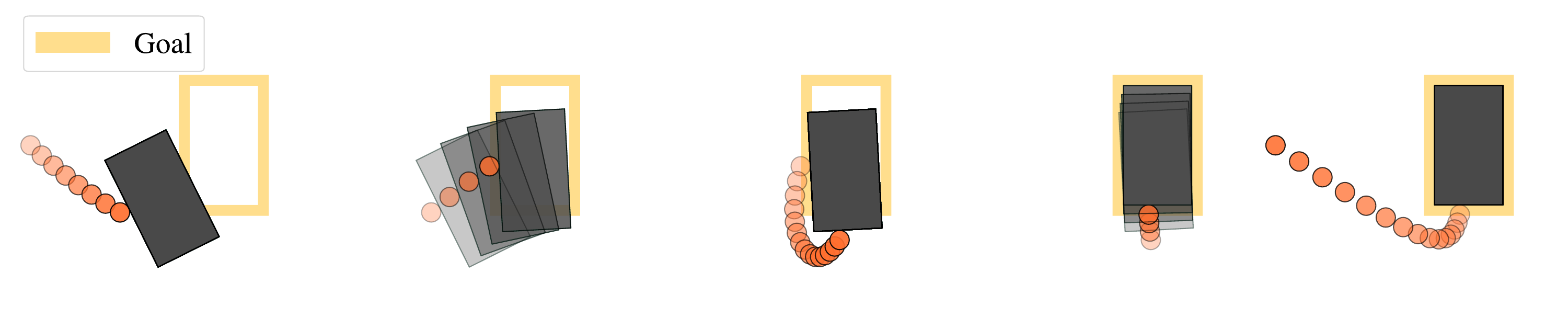}
       \caption{Our method}
    \end{subfigure}
	%\begin{subfigure}[b]{0.8\linewidth}
	%		\centering
	%		\includegraphics[width=\linewidth]{media/direct_traj_trajectory.pdf}
 %      \caption{The baseline}
 %   \end{subfigure}
	\begin{subfigure}[b]{0.8\linewidth}
			\centering
			\includegraphics[trim=0 0 0 10pt, clip, width=\linewidth]{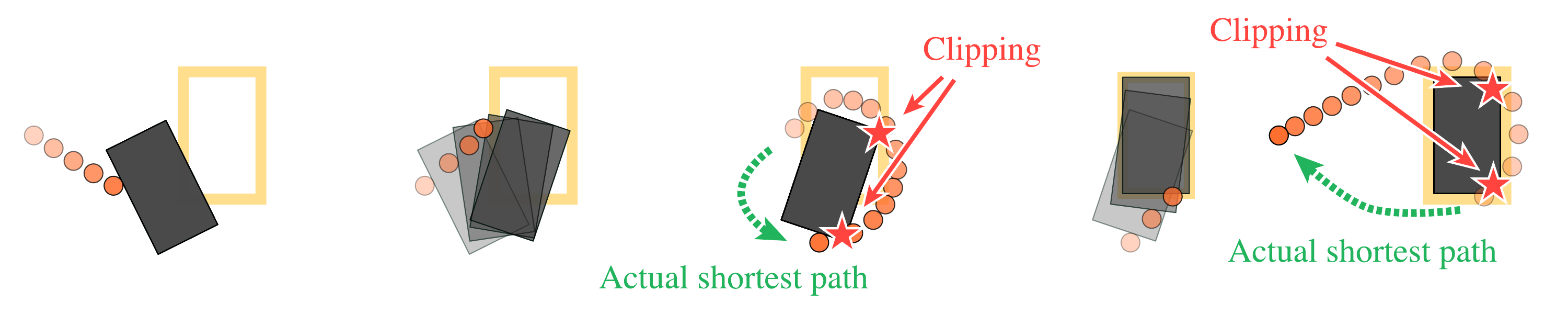}
       \caption{Contact-implicit method}
    \end{subfigure}
   \caption{A comparative example between our method and a contact-implicit method. Our method picks the shortest path around the object, while the baseline goes the longer way around twice, highlighting the fact that our method is capable of global reasoning. Our method also guarantees that the trajectory stays collision-free between contacts, while the baseline can be seen to clip the corners of the slider.
   }
   \label{fig:comparison}
\end{figure*}

This section contains both numerical and hardware experiments. 
An open-source implementation of the proposed motion planning algorithm, as well as the code used to run the experiments, can be found at \url{https://bernhardgraesdal.com/rss24-towards-tight-convex-relaxations/}.

We show an example of a generated motion plan in \Cref{fig:plan_example}.
We choose the cost weights for contact modes to $k_{p^P} = k_{p^S} = k_{v^P} = 10$, $k_{v^S} = 100$, $k_f = 10$, $k_T = 1$, and $k_\phi = 0.1$ for the non-contact modes, where the only tuning that has been done is choosing reasonable orders of magnitude for the cost parameters, ensuring that the cost terms contribute approximately equally to the cost.
We consider two different slider geometries, one simple box-shaped geometry and a T-shaped geometry, with the \acrshort*{com} located according to a uniform mass distribution.
For reference, both slider geometries have a maximum "radius" close to $0.25$ meters, and the plans are generated within a box with sides $0.6$ meters.

\subsection{Planner performance}
We evaluate the global optimality of the motion planner by generating 100 motion plans for the two slider geometries, with random initial and target configurations drawn from an uniform distribution, and upper-bound their respective optimality gaps according to~\eqref{eq:optimality_bound}.
We use the conic solver Mosek \cite{mosek} to solve the optimization problem on a laptop with an Apple M1 Max chip with 32GB of RAM.
For both slider geometries, we achieve a success rate of $100 \%$, that is, the rounding step is able to retrieve a feasible solution for all the generated problem instances.
We show the average planning times, average rounding time, and average optimality gaps in~\Cref{tab:plan_generation}.
We achieve a mean upper bound on the optimality gap of around $10\%$, with a median optimality gap of less than $8\%$ for both slider geometries.
We generate plans in a few seconds for the box-shaped slider, and in approximately $1.5$ minutes for the T-shaped slider.
The mean solve time for the box-shaped slider is an order of magnitude lower than for the T-shaped slider, which has twice as many faces, and hence generates a much bigger \acrshort*{gcs} problem.

%!TEX root = ../root.tex
\begin{table}[htb]
	\centering
	\begin{tabular}{|c|c|c|c|c|} \hline
		Slider & \acrshort*{sdp} solve time & Rounding time & Optimality gap ($\delta_\text{round}$) \\
  \hline
    Box & 7.05s (6.87s) & 0.05s (0.05)s& 8.33\% (5.39\%) \\
    Tee & 83.61s (80.12) & 0.36s (0.014s) s& 10.41\% (7.47\%) \\
  \hline
	\end{tabular}
	\caption{
    Solve times, rounding times, and upper bounds on the global optimality gaps for planning planar pushing tasks with two different slider geometries, for 100 randomly chosen initial and final configurations. The box-shaped slider has 4 faces and the T-shaped slider has 8 faces. The reported values are mean values, with the median values shown in parenthesis.}
     \label{tab:plan_generation}
\end{table}

Next, we show statistics for the generated convex optimization problems for the two different slider geometries in~\Cref{tab:problem_stats}.
It is clear that the size of the optimization problem grows quickly (although polynomially) with the number of faces on the slider.
We believe that we can significantly reduce the size of the optimization problem and thus reduce solve times by better exploiting the heavily structured problem formulation, but leave this for future work.

\begin{table}[htb]
	\centering
    \begin{tabular}{|c|c|c|}
        \hline
        Problem statistics & Box & Tee \\
        \hline
        Number of constraints & 48 846 & 212 795\\
        Number of scalar variables & 8 854 & 39 078 \\
        Number of \acrshort*{psd} matrices & 88 & 368  \\
        \hline
    \end{tabular}
    \caption{Statistics for the convex optimization problem that our methods formulates. The size of each \acrshort*{psd} matrix is $15\times 15$, and $3$ knot points are used for each mode.}
    \label{tab:problem_stats}
\end{table}

\subsection{Comparison with contact-implicit trajectory optimization}
\label{sec:comparison}
To compare our method with a state-of-the-art baseline for contact-rich planning, we select a direct, contact-implicit trajectory optimization method similar to those proposed in~\cite{posaDirectMethodTrajectory2014} and~\cite{manchester2020variational}.
This method encodes contact modes implicitly using complementarity constraints that are relaxed and solves the problem as a sequence of \acrshort*{nlps} with increasingly strict complementarity constraints until they are satisfied with equality.
This method requires an initial guess, which we provide as a straight-line interpolation between the initial and target configuration.
We generate another $100$ random initial and target configurations from an uniform distribution, and solve the planning problems with both methods.

%To fairly compare the quality of the generated plans, we minimize only the total arc length for the pusher and slider.
%This is because the baseline uses a fixed trajectory duration while our method is able to dynamically change the length of the trajectory to minimize the cost, and hence any comparison that relies on a cost that is dependent on time (including velocity costs) will be dependent on the pre-specified trajectory duration and thus unfair.

\begin{table}[htb]
	\centering
    \begin{tabular}{c|c|c|}
        \cline{2-3}
        & \multicolumn{2}{c|}{Success Rate} \\
        \hline
        \multicolumn{1}{|c|}{Slider} & Our method & Contact-implicit method\\
        \hline
        \multicolumn{1}{|c|}{Box} & 100\% & 58\%\\
        \multicolumn{1}{|c|}{Tee} & 100\% & 12\% \\
        \hline
    \end{tabular}
    \caption{A comparison of the success rate between our method and a contact-implicit trajectory optimization method. As our method is capable of global reasoning and does not rely on an initial guess, it has a much higher success rate compared to the baseline.}
    \label{tab:comparison}
\end{table}

\Cref{tab:comparison} presents a comparison between our method and the contact-implicit method.
With a cost function that is specifically tuned to optimize the baseline's performance, our method finds a solution in $100\%$ of the instances for both slider geometries. In contrast, the baseline often fails, finding a solution in $58\%$ of the instances for the box-shaped slider geometry and a mere $12\%$ for the T-shaped slider.
For the T-shaped slider, the contact-implicit method appears to succeed only when the initial and target configurations require a small number of mode switches that are relatively close to the initial guess. 
This limitation is not surprising, as the baseline is a local method that relies heavily on its initial guess.
The T-shaped slider has a significantly higher number of possible mode sequences and a more complex \acrshort*{sdf} compared to the box-shaped geometry, making it more difficult for the baseline to find a solution.
On the other hand, our method solves all instances for both geometries. This highlights a key advantage of our approach: by reasoning on a global level, our method (empirically) always finds a solution, without relying on an initial guess.

%The two transcriptions implement the cost function differently and rely on distinct pre-specified parameters, such as the number of knot points and trajectory/mode durations. These differences make it challenging to conduct a rigorous comparison of the plan quality between the two methods.
The two transcriptions rely on different pre-specified parameters, such as the number of knot points and trajectory/mode durations. These differences make it challenging to conduct a rigorous comparison of the plan quality between the two methods.
Instead, we provide a qualitative analysis to highlight some advantages of our method.
We start by pointing out that the performance of the contact-implicit method in~\cref{tab:comparison} required significant tuning of the cost function and problem numerics.
After this tuning however, the found plans empirically seem very reasonable. 
In contrast, our method requires minimal tuning and no initial guess while still finding high-quality plans for all given problem instances.
Our method also provides a good metric for evaluating solution quality through the obtained upper bound on the optimality gap, unlike the baseline, where such an objective comparison is difficult.

Furthermore, the baseline requires careful consideration to avoid problems with unwanted contacts and penetration between discretization points.
While these problems can certainly be addressed by e.g. finer trajectory discretization, velocity constraints, or tuning, these issues are naturally handled in our method, which guarantees that the entire path is collision-free between contacts, independently of the aforementioned factors.
Additionally, while the baseline requires a fixed, pre-specified trajectory duration, our method can vary the trajectory length through the number of visited contact modes.
However, it is important to note that our method still relies on a pre-specified trajectory duration in each contact mode.
\Cref{fig:comparison} illustrates some of the mentioned qualitative differences through a comparison of trajectories generated by the two methods.

\subsection{Execution on real hardware}
Finally, we demonstrate the feasibility of the obtained motion plans on a Kuka LBR iiwa 7 R800 7-DOF robotic arm, with a T-shaped slider object.
A picture of the experimental setup is shown in \Cref{fig:experimental_setup}.
We experimentally determine approximate values for the friction parameters from a handful of open-loop executions, and find the friction coefficient and integration constant for the contact between the slider and table in~\eqref{eq:limit_surface_const} to be \(\mu_S = 0.5\) and \(c = 0.3\), and the friction coefficient between the pusher and slider in~\eqref{eq:friction_cone} to be \(\mu = 0.05\).
We find that the plans generated with these friction parameters often perform well open-loop. For extra stability, we use a feedback controller to execute the plans on hardware, and employ 
a hybrid \acrfull*{mpc} commonly used for pushing tasks \cite{hoganFeedbackControlPusherSlider2020,hoganReactivePlanarManipulation2018}.
Two different plans and the corresponding hardware execution can be seen in \Cref{fig:experiments_and_plans}, and a collection of hardware demonstrations can be seen in the video at
\url{https://bernhardgraesdal.com/rss24-towards-tight-convex-relaxations/}.

\section{Conclusion and Future Work}
\label{sec:conclusion}
%!TEX root = ../root.tex

In this work, we present a framework for planning near-globally optimal trajectories for contact-rich systems.
We demonstrate how the dynamics of a contact-rich system can be described as a \acrshort*{qcqp}, and leverage semidefinite relaxations to approximate the planning problem in a fixed contact mode as a convex program.
This convex approximation of the dynamics enables us to plan complex trajectories over mode sequences by leveraging the \acrshort*{spp} in \acrshort*{gcs} framework from \cite{marcucci2023shortest}.
The entire planning problem can therefore be solved as a single convex optimization followed by a cheap rounding strategy based on local optimization.
As an initial application, we study our framework on the planar pushing problem and demonstrate that our method empirically generates near-globally optimal solutions. We are currently working on applying our method to more complex contact tasks.

While the proposed approach is more efficient than prior global optimization methods, the planning times are still too slow for fast, reactive planning.
We are currently working on better exploiting our highly structured problem formulation,
both through the use of a custom solver,
more compact problem formulations,
and more efficient construction of the underlying optimization problem.
Additionally, there is a rich literature on both exploiting special structure particularly in \acrshort*{sdp}s \cite{garstkaCliqueGraphBased2020, burer2003nonlinear}, as well as approximating \acrshort*{sdp}s using simpler cones \cite{majumdar2020recent, blekherman2022sparse} which could potentially reduce solve times by several orders of magnitude.
Future work will explore the ability of these reduction methods to accelerate the planning.
In addition, we are exploring how further techniques from the literature on \acrshort*{sdp} relaxations can be used to automatically provide even tighter relaxations in a reasonable computational budget.

\section*{Acknowledgement}
\label{sec:acknowledgement}
%!TEX root = ../root.tex

This research was supported by (in alphabetical order):
Aker Scholarship;
Amazon.com Services LLC, PO No. 2D-12585006;
Boston Dynamics AI Institute; and 
Office of Naval Research, Award No. N00014-22-1-2121.
Any opinions, findings, conclusions, or recommendations
expressed in this material are those of the authors and do not necessarily reflect the views of the
funding agencies.

\bibliographystyle{IEEEtran}
\bibliography{IEEEabrv,ref}

\end{document}